\newdimen\bibsep
\documentclass[preprint,5p,nonatbib]{elsarticle}
\let\bibsep\relax

\usepackage[T1]{fontenc}
\usepackage[utf8]{inputenc}
\usepackage{newtxtext,newtxmath}  
\usepackage{microtype}
\usepackage{graphicx}
\usepackage{booktabs}
\usepackage{tabularx}
\usepackage{enumitem}
\setlist{itemsep=2pt,topsep=4pt}

\usepackage[natbibapa]{apacite}
\makeatletter\NAT@longnamesfalse\makeatother  

\AtBeginDocument{%

  \urlstyle{same}%
}

\usepackage[colorlinks=true,allcolors=blue!50!black]{hyperref}

\AtBeginDocument{\renewcommand{\doi}[1]{\url{https://doi.org/#1}}}
\journal{Computers and Education: Artificial Intelligence}

\begin{document}

\begin{frontmatter}

\title{PRAISE: A Reproducible Workflow for Non-Programmer Scientists to Implement Published AI Models}

\author[1,2]{Max Achterweust}
\author[1,3]{Stefan Groot Nibbelink}
\author[1,4,5]{Vincent J. de Beer}
\author[2]{Stefan Schipper}
\author[2]{Zeynep Y\"uksektepe}
\author[2]{Pedro Andrade da Silva}
\author[2]{Aurelia Peppelaar}
\author[1,5,6]{Martijn J. van der Willigen}
\author[2]{Christophe B. Minkenberg\corref{cor1}}
\ead{c.b.minkenberg@hr.nl}
\author[1,5,7]{Robert F. van der Willigen}

\cortext[cor1]{Corresponding author}

\affiliation[1]{organization={HR Datalab EAS, School of Engineering and Applied Sciences, Rotterdam University of Applied Sciences},
  city={Rotterdam}, country={the Netherlands}}
\affiliation[2]{organization={Department of Chemistry, School of Engineering and Applied Sciences, Rotterdam University of Applied Sciences},
  city={Rotterdam}, country={the Netherlands}}
\affiliation[3]{organization={Department of Electrical Engineering, School of Engineering and Applied Sciences, Rotterdam University of Applied Sciences},
  city={Rotterdam}, country={the Netherlands}}
\affiliation[4]{organization={Department of Biology and Medical Laboratory Research, School of Engineering and Applied Sciences, Rotterdam University of Applied Sciences},
  city={Rotterdam}, country={the Netherlands}}
\affiliation[5]{organization={Data Supported Healthcare: Data-Science Unit, Research Center Innovations in Care, Rotterdam University of Applied Sciences},
  city={Rotterdam}, country={the Netherlands}}
\affiliation[6]{organization={School of Communication, Media and Information Technology, Rotterdam University of Applied Sciences},
  city={Rotterdam}, country={the Netherlands}}
\affiliation[7]{organization={HR DataLab AI SusTech, Centre of Expertise HRTech, Rotterdam University of Applied Sciences},
  city={Rotterdam}, country={the Netherlands}}

\begin{abstract}
Artificial Intelligence (AI) models are increasingly developed, published and used across scientific fields. Researchers, students and industry professionals who want to apply these models to their own research questions often lack the programming and data science expertise to do so. Existing guidelines cover single steps, but not the route from a published model to working predictions on one's own data. Here we present the hands-on workflow PRAISE, a Practical, Reproducible AI-model Implementation for Science and Education, that helps non-programmers implement published AI models step by step. Because implementations rarely work the first time, PRAISE anticipates common failures and provides a fallback route for each. As a proof of principle, three BSc chemistry students and their tutor applied PRAISE without prior programming or data science knowledge. They reproduced, validated and extended a state-of-the-art model that predicts enzyme-substrate interactions. This shows that with PRAISE, implementing an AI model from the literature becomes a learnable skill, one that turns non-programmer scientists from users into evaluators and improvers of AI models.
\end{abstract}


\end{frontmatter}

\section{Introduction}\label{sec:introduction}

Artificial intelligence (AI) has emerged as a rapidly advancing field, generating an increasing number of applications with societal relevance \citep{Jumper2021,Yang2025,Sajadieh2026}. Researchers, students and industry professionals, e.g. in STEM disciplines, who want to apply published AI models to a problem often encounter difficulties implementing and operationalizing these models. Implementing a published AI model in a new setting may be difficult for several reasons. Professionals may lack data-science expertise or suitable computational infrastructure. They may also be unfamiliar with the installation, training, and validation of AI pipelines. Insufficient AI literacy can further discourage professionals from initiating the implementation process \citep{Biagini2025,Pinski2024}. Bridging the gap between the conventional scientific disciplines and data science is essential for useful and ethical AI model implementation \citep{PionTonachini2021}.

This paper presents PRAISE, a Practical, Reproducible AI-model Implementation for Science and Education, a hands-on workflow that helps non-programmer scientists with a reproducible implementation of a published AI model to their own practical application. Existing AI lifecycles and workflows describe the pipeline design, data preparation, training, validation, and deployment \citep{ISO2022b,Amershi2019,Kreuzberger2023,OECD2019,NIST2023}, however these workflows stop short of the practical requirements for implementing an existing, literature-derived model. A different body of work addresses non-programmer scientists directly, through reporting guidelines covering data curation, representation, training, and reproducibility \citep{Artrith2021,Walsh2021}, and low-code environments for exploring machine learning (ML) \citep{Magnano2022}. These resources each address a particular part of the process. The route from a published model to a working deployment is left undescribed.

To avoid any possible confusion, we fix some terminology here that is used throughout this work: implementation is the umbrella term for the whole route of selecting, installing, running and validating an AI model. Reproduction indicates running the published model and obtaining metrics comparable to those reported. Adaptation means retraining the model on data the user adds, and/or with changes in the model pipeline. Deployment, the final stage of the AI life cycle \citep{ISO2022b,OECD2019}, denotes making the validated model available for new predictions in the use case, e.g. as a script or web interface.

The paper is structured as follows: Sections~\ref{sec:how-an-ai-model-learns-from-data-makes-predictions-and-is-validated} and \ref{sec:computational-stack-and-development-tools-for-ai-models} supply background on how an AI model works, and the layered computing infrastructure it runs on, including the role of generative AI as a coding aid. Look them up whenever a step is unclear. The PRAISE workflow, from use case formulation to deployment, is set out in Section~\ref{sec:praise-the-workflow-for-the-implementation-of-published-ai-models}. Section~\ref{sec:fallback-troubleshooting-by-workflow-iteration} describes common issues and points to fallback loops in the workflow that solve them. Section~\ref{sec:case-study-of-workflow-application-in-research-and-education} follows the workflow through a use case carried out by non-programmers.

Two strategies are suggested how this paper may be read: readers seeking a systematic understanding can follow the sections in sequence. Those who want to start implementing a model directly can begin in Section~\ref{sec:praise-the-workflow-for-the-implementation-of-published-ai-models} and use the preceding two sections as reference material.

\section{How an AI model learns from data, makes predictions and is validated}\label{sec:how-an-ai-model-learns-from-data-makes-predictions-and-is-validated}

A predictive AI model learns relationships between input data and known outcomes during training, subsequently the model uses these learned relationships to make predictions for new data. The validity of this process depends on the quality of the input data and how the neural network is trained. Evaluation is essential to safeguard the validity. In this section the data, training process and validation of the outcomes are discussed.

\subsection{Data as the basis for learning}\label{sec:data-as-the-basis-for-learning}

AI models need to be trained on relevant, fair and valid data. For training to be successful, the dataset must fulfill several requirements.

\subsubsection{Dataset size}\label{sec:dataset-size}

The dataset should be of sufficient size to represent the underlying population. Small datasets may lead to overfitting, reducing the reliability of predictions \citep{Rajput2023}. Model performance generally improves with increasing dataset size until a saturation point is reached \citep{Rajput2023}. What constitutes a `small' or `sufficiently large' dataset depends on data complexity, model architecture, and task, since learning is more difficult when input data is highly variable or unstructured \citep{Naser2026}. As an empirical rule, a minimum number of samples per feature is often given, from 3 to 300 samples per feature, dependent on the intended use \citep{Naser2026}.

\subsubsection{Non-ideal dataset properties}\label{sec:non-ideal-dataset-properties}

Datasets may suffer from imperfections that affect the applicability and reliability of the resulting AI models. Non-ideal data distributions take several forms:

Gaps in training data reduce the applicability and reliability of a model. For example, an AI model trained to predict chemical toxicity may perform poorly when certain categories of compounds are underrepresented or absent from the dataset. Such gaps reduce the applicability domain of the model and limit its practical relevance \citep{Hartung2025}.

Biased datasets contain features or categories that are disproportionately represented relative to the true underlying population \citep{Eid2021}. In a toxicology context, a dataset dominated by molecules with similar physicochemical properties may unintentionally specialize the model toward those compounds, causing performance to deteriorate for chemically distinct structures. \citet{ChenBao2025} intentionally created biases in a dataset about strength of concrete beams and found that this compromised ML model accuracy and universality.

Skewness is an asymmetry in the data distribution \citep{Monard2002}. In the case of the toxicology example, a dataset containing 99\% non-toxic chemicals and only 1\% toxic compounds is skewed towards the majority class. Such imbalance is not necessarily unrealistic. Many real-world datasets, such as fraudulent transaction detection datasets, naturally contain rare minority classes \citep{Monard2002}. Increasing the dataset size will disproportionately benefit minority class representation \citep{Weiss2001}. Model performance increased when the minority classes were represented in the dataset for 50\% instead of their true 19\% presence \citep{Weiss2001}.

A related issue is the absence of confirmed negative datapoints. In many datasets, negatives are constructed by random negative sampling, which can introduce false negatives into both training and validation sets \citep{Eid2021}.

\subsubsection{Dataset splitting}\label{sec:dataset-splitting}

Before usage, the data are generally split into separate subsets for testing, training and validation. This separation is important because the model should be evaluated on data that were not used to adjust its parameters. Conventional partitions are an 80:20 ratio for a two-way train-test split and 64:16:20 for a train-validation-test split, although no consensus exists on these proportions and an optimal split depends on the number of parameters in a linear regression model \citep{Joseph2022}. The partition itself is often generated randomly and is therefore controlled by a random seed. Reproducible splits can be made by recording and reporting the seed \citep{Wegmeth2023,Lyu2021}.

Data leakage occurs when a model is trained on information it should not have access to during training, often resulting in unrealistically high model performance \citep{Apicella2025}. The authors propose a data-leakage taxonomy, based on the place in the pipeline where the data leakage occurs. In Section~\ref{sec:dataset-preparation} a leakage protocol checklist is proposed.

\subsection{From input data to a prediction}\label{sec:from-input-data-to-a-prediction}

Once a dataset has been assembled and split, the model itself determines how the input variables are converted into a prediction. This section describes the structure of a neural network, how it learns through minimizing the loss function, and which types of machine learning exist.

\subsubsection{Structure of neural networks}\label{sec:structure-of-neural-networks}

A neural network consists of a chain of layers of nodes, sometimes referred to as neurons. Besides the input and output layers, the network may consist of several hidden layers in the case of deep learning. The length of the chain gives the network its depth.

Information is fed to the nodes of the input layer and then moves from one layer to the next, until it reaches the output layer, which gives the results of the network. Each layer receives the previous layer's outputs as inputs and passes its own outputs to the subsequent layer. In this way, each consecutive hidden layer applies a further transformation to the representation produced by the preceding layers. This enables the network to transform input data into a representation that facilitates modelling, while optimizing its parameters to capture the underlying relationships accurately \citep{Goodfellow2016a}. Each node multiplies its inputs by weights, sums them, adds a bias, and passes the result through a nonlinear activation function to produce its output \citep{Goodfellow2016a}. This nonlinearity allows the network to capture complex relationships in the data.

\subsubsection{Loss function minimization}\label{sec:loss-function-minimization}

The structure of a neural network, i.e. the number of layers, the number of nodes per layer and the activation functions, is typically fixed from the outset, while the weights and biases are optimized during training. In supervised learning, a loss function quantifies the discrepancy between a neural network's prediction and the corresponding target value, thereby measuring prediction error. A fully trained AI model has weights and biases such that the loss function is minimal on the training set. An optimal set of these parameters is obtained during the training of the network.

At the start of a training session, some typical values for these parameters are chosen (randomly or with some background knowledge). A single batch of the training dataset is then fed to the neural network and the output is determined.

During backpropagation, the prediction error is propagated backward through the network. Gradients are calculated for each weight and bias, and an optimization algorithm uses them to update these parameters and reduce the loss \citep{Narihira2021,Goodfellow2016a}. The updated parameters are then used in the next forward pass of the following batch. This process of calculating a prediction, calculating the loss, determining the gradients and updating the parameters is repeated batch by batch until the process has passed the whole training dataset. This completes one epoch of the training cycle. The number of batches per epoch depends on the selected batch size.

Full training of an AI model often involves many epochs. The loss should generally decrease over successive training epochs, although fluctuations may occur, particularly during the initial epochs \citep{Fort2020}.

\subsubsection{Loss function monitoring and early stopping}\label{sec:loss-function-monitoring-and-early-stopping}

To determine when an AI model is well trained, the loss function should be monitored on both the training set and a held-out validation set. When the training loss (loss function evaluated on the training dataset) continues to decrease while the validation loss (loss function evaluated on the validation dataset) starts to increase, this indicates that the model is overfitting the training data. The minimum validation loss can therefore be used to identify an appropriate point for stopping training, often using an early stopping procedure \citep{Prechelt2012}. The weights obtained at the selected point during training are subsequently used to calculate predictions for new, previously unseen input data.

\subsubsection{Common types of machine learning}\label{sec:common-types-of-machine-learning}

What exactly a network learns depends on the type of learning that is being employed. In supervised learning every input is accompanied by a known true value, and the network learns to produce a correct output for inputs it has not seen \citep{Goodfellow2016b}. The loss function in this case measures the error between the output scores and the desired scores, and learning adjusts the weights to reduce it \citep{LeCun2015}.

In unsupervised learning only the inputs are available, and the network learns properties of the data itself, for example to group entries by their distinct features \citep{Goodfellow2016b}. The loss function in unsupervised learning expresses how well the model captures the underlying data patterns \citep{Goodfellow2016b}.

In reinforcement learning, a neural network observes a connected environment that gives an input, after which the neural network chooses an action that changes the environment and results in a new input for the network \citep{Kaelbling1996}. In a guided trial and error effort the neural network can learn from its actions and their consequences on the environment \citep{Kaelbling1996}.

\subsection{Verifying model training and predictive performance}\label{sec:verifying-model-training-and-predictive-performance}

Training diagnostics assess whether the model was trained as intended and whether the training process shows indications of failure. This includes checking whether the training was completed successfully, if the loss function decreased during the epochs in the training, and how the learned parameters behave during the training.

Weight distribution analysis indicates how weights behave during training. Adjusting the weights and assessing the outcome is how a model learns, and weights represent the strength and direction of relationships between neurons within the network, e.g. a negative number meaning a negative relationship \citep{Goodfellow2016a}. In weight distribution analysis, it is examined how abstract parameters relate to each other during or after training \citep{Glorot2010}. These weights are visualized for one layer or for the whole network and plotted as a histogram \citep{Li2024,Glorot2010}. The weights are plotted on the x-axis. The resulting distributions can be compared between training stages or between different training runs \citep{Li2024,Glorot2010}.

Most measures of classification performance derive from the confusion matrix, which counts true positives, true negatives, false positives and false negatives. Accuracy expresses the proportion of all predictions that are correct, and precision expresses the proportion of predicted positives that are truly positive \citep{Hicks2022}. Accuracy can be misleading when classes are imbalanced, because a model can achieve high accuracy by predominantly predicting the majority class while performing poorly on the minority class. Precision should be interpreted together with recall (sensitivity), which measures how many of the actual positive cases are detected \citep{Hicks2022}.

After training, model performance should be validated against known outcomes using statistical metrics. For binary classifiers, the Receiver Operating Characteristic (ROC) curve plots the true positive rate against the false positive rate across all decision thresholds. The area under this curve (ROC-AUC) provides a threshold-independent measure of predictive capability: values near 1 indicate strong performance, 0.5 reflects random classification, and values below 0.5 indicate performance worse than random ranking \citep{Gneiting2022}.

ROC-AUC can be a useful measure, but does not directly reflect the precision of positive predictions and can therefore provide an incomplete picture when performance on the minority class is of particular interest \citep{Davis2006,Powers2011}. For example, in a model trained to detect a rare disease within a largely healthy population, poor detection of the minority class may have only limited influence on the overall ROC curve despite being practically important.

For imbalanced datasets, the Matthews Correlation Coefficient (MCC) is often more informative than the ROC-AUC. MCC incorporates all four outcomes of the confusion matrix \citep{Chicco2023}. The metric ranges from $-1$, representing complete disagreement between prediction and observation, to 1, representing perfect classification; a value around 0 indicates performance no better than random prediction. Because MCC considers all classes simultaneously, it is regarded as a balanced performance metric for skewed datasets. Another validation method for imbalanced datasets is the precision-recall curve (PRC) \citep{Saito2015}.

Data leakage should always be checked to ensure honest model performance. This can be assessed by checking whether the training and test datasets are clearly separated, whether duplicate or highly similar samples occur in both datasets, and whether preprocessing steps such as normalization, feature selection, or data augmentation were performed using the complete dataset before splitting \citep{Kapoor2024,Kapoor2023}. All are potential causes of data leakage.

\section{Computational stack and development tools for AI models}\label{sec:computational-stack-and-development-tools-for-ai-models}

\begin{figure*}[tbp]
  \centering
  \includegraphics[width=.75\textwidth]{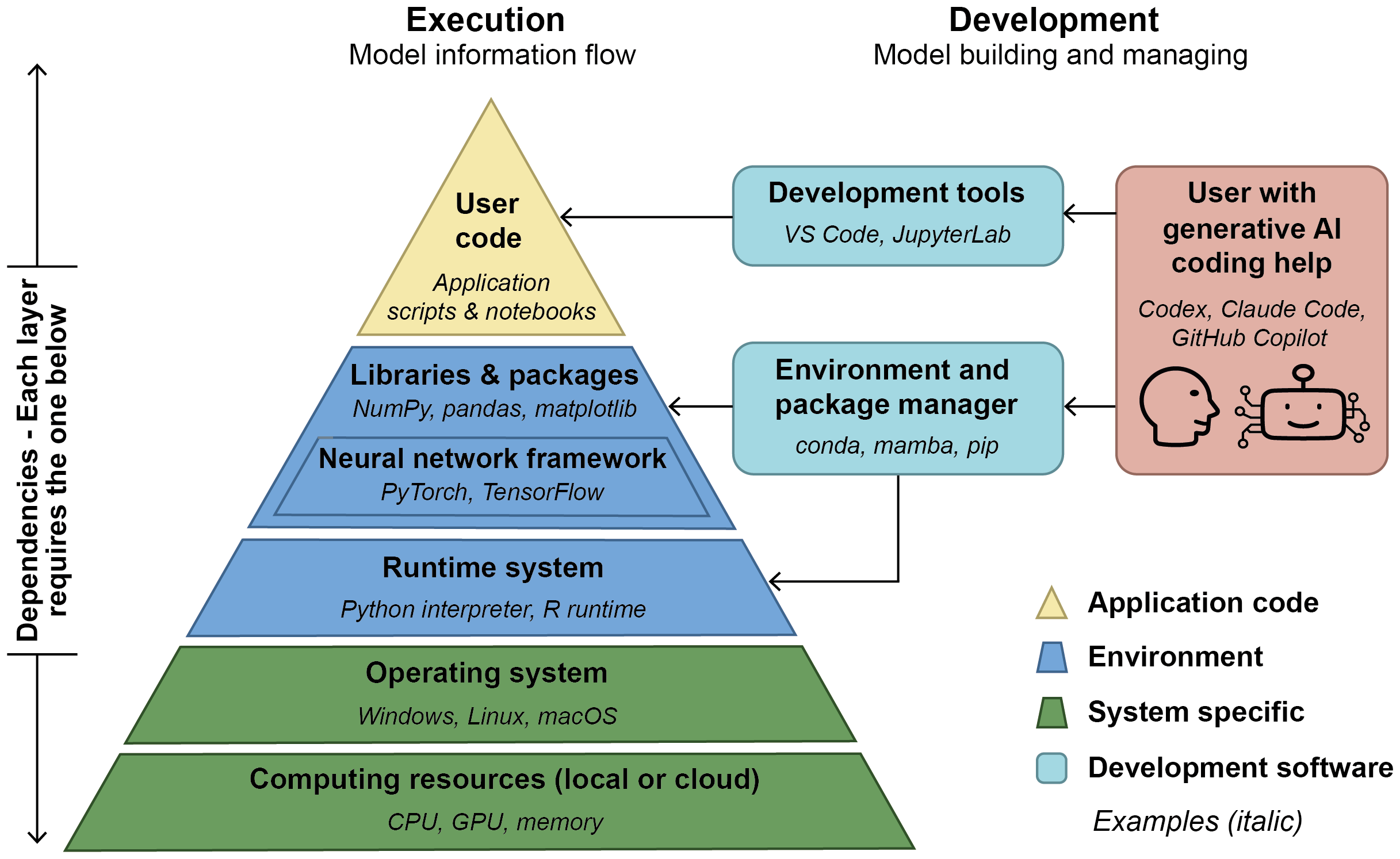}
  \caption{Visualization of the computational stack required to execute an AI model (left), and the tools that configure the stack (right). During a run, information flows from the top to the bottom of the stack and back. All layers and dependencies must be installed and compatible before the AI model can run. Users primarily work in the development tool and in the environment and package manager. Generative AI (GenAI) can help the user to develop the model and stack.}
  \label{fig:stack}
\end{figure*}

An AI model runs on a layered stack of hardware and software. Each layer of software depends on the layer below it, so each needs to be present in a compatible version. In this stack, information of the running AI model flows from top to bottom and back. An environment is made, which is a project-specific collection of a programming language interpreter and installable packages. The stack is built and externally maintained, through development tools and an environment manager that the user operates (increasingly with generative AI as coding assistance). Figure~\ref{fig:stack} represents this structure schematically, with the execution path on the left and the developmental path on the right. Each layer is discussed in turn, from the bottom up.

\subsection{Computing resources}\label{sec:computing-resources}

Computing resources are hardware that comprise the physical components to perform computations. Given the computational demands of AI, access to a capable GPU is strongly recommended. GPUs are optimized for parallel processing, enabling efficient handling of large, high-dimensional datasets typical in machine learning workflows \citep{Raina2009}.

\subsubsection{Local hardware}\label{sec:local-hardware}

High-performance or gaming PCs are decent entry-level AI hardware, especially when the system contains a dedicated GPU. Memory is often a bottleneck in training complex AI models, and ways to decrease memory consumption during training can be found in Section~\ref{sec:revising-user-code} \citep{Chen2016,Ojika2020,Rajbhandari2020}.

\subsubsection{Cloud computing resources}\label{sec:cloud-computing-resources}

An alternative to local hardware is cloud computing. Cloud computing is the on-demand use of scalable off-site computing resources \citep{ISO2023}. Commercial cloud providers for AI applications include Microsoft Azure, Amazon Web Services (AWS), Google Cloud Platform (GCP) and Oracle Cloud Infrastructure (OCI). Costs depend on factors such as computing power, user agreement duration, and the use of spot instances. Spot instances are discounted computing capacity offered when the cloud service provider has overcapacity \citep{Amazon2026}. National research and education networks (NRENs) are government-funded academic cloud infrastructure, such as the Dutch IT cooperative Samenwerkende Universitaire RekenFaciliteiten (SURF), and may offer a cost-effective and safer alternative for eligible researchers \citep{SURF2026a,GEANT}.

A practical concern with commercial cloud computing is data security. Sensitive data uploaded to off-site infrastructure may be subject to cross-border legal frameworks. For example, U.S. legislation can compel disclosure of data held by U.S.-based providers, even when stored in Europe \citep{DOJ2018}. Cloud providers may also use uploaded data for service improvement or AI training purposes, although opt-out mechanisms are available. It is advised not to upload privacy-sensitive information to off-site infrastructure, such as a cloud provider, but to anonymize sensitive data or work with synthetic data to reduce privacy risks \citep{vanVelzen2025,EU2016}.

\subsubsection{Hybrid computing infrastructure}\label{sec:hybrid-computing-infrastructure}

A hybrid infrastructure combining local and cloud computing may provide a practical solution for larger AI workloads. Initial installation, troubleshooting, and exploratory runs can first be performed locally to familiarize users with the model and identify technical issues before migrating to cloud infrastructure. This approach reduces unnecessary cloud GPU usage and associated costs in the startup phase.

\subsection{Operating system}\label{sec:operating-system}

The computing hardware is controlled by its operating system (OS), which is called system software. Examples are Linux, Windows and macOS \citep{Wang2023}. Compatibility between AI models and OS environments is not guaranteed; even minor version differences can introduce instability or errors \citep{Wang2023}. Reproducing the original development environment as closely as possible is therefore advisable. This can be done via a local container, such as a Linux container \citep{Flinders2025}.

\subsection{Runtime system}\label{sec:runtime-system}

The operating system interacts with a user's applications via a runtime system. The runtime system is software that allows code to execute, which is therefore programming language specific. Python, for example, is a computer language with its own interpreters, like CPython or PyPy \citep{PSF}. The interpreter's task is to translate the programming language like Python in order for the operating system to understand it, perform tasks and allocate memory \citep{Aho2006,PSF}. For GPU-accelerated workloads, additional runtime components are often required. A common example is CUDA, which provides the necessary interface between software and NVIDIA GPUs, handling memory allocation and parallel task execution.

\subsection{Libraries and packages}\label{sec:libraries-and-packages}

Above the runtime system sits a layer of libraries and packages. This is reusable code that the application code calls upon for tasks it does not perform itself. A library is a collection of functions written for a general purpose, and a package is the distributable unit in which a library is installed. In Python, pandas provides functions for reading and manipulating tabular data such as CSV files \citep{McKinney2010}, NumPy provides array operations \citep{Harris2020}, and matplotlib renders results as graphs \citep{Hunter2007}. Libraries of this kind support loading data, transforming it, and visualizing outcomes. The application code lists them as dependencies, and each one must be present in the environment in a compatible version.

\subsubsection{Neural network frameworks}\label{sec:neural-network-frameworks}

Neural network frameworks are libraries that form the executional backbone of AI models. These frameworks consist of three components: variables, functions, and parametric functions \citep{Narihira2021}. Variables represent the data used in the model and are stored in multidimensional arrays, such as vectors or tensors \citep{Narihira2021}. Functions perform mathematical operations on these variables, whereas parametric functions contain trainable parameters that are adjusted during training according to relationships identified in the training data. The model's weights are stored within these parametric functions. The building of a model with parametric functions is more straightforward thanks to neural network frameworks like TensorFlow \citep{Abadi2016} and PyTorch \citep{Paszke2019}, because they automatically track operations in the forward pass and compute gradients in the backward pass \citep{BaAlawi2025}.

\subsection{User code: application scripts and notebooks}\label{sec:user-code-application-scripts-and-notebooks}

The top layer holds the code the user writes. In a single script or notebook, the user calls on the packages from the layers below and arranges them in the order the work requires: loading the data, preparing it with the appropriate libraries, and passing it to the neural network framework in small portions called batches. The same file specifies how the model is trained, which settings the optimization uses, and at which points the results are checked against validation data. Once the model has produced its predictions, supporting libraries turn these into figures and tables the user can interpret. The user code can be executed as a script in the used development tool.

\begin{figure*}[t]
  \centering
  \includegraphics[width=.75\textwidth]{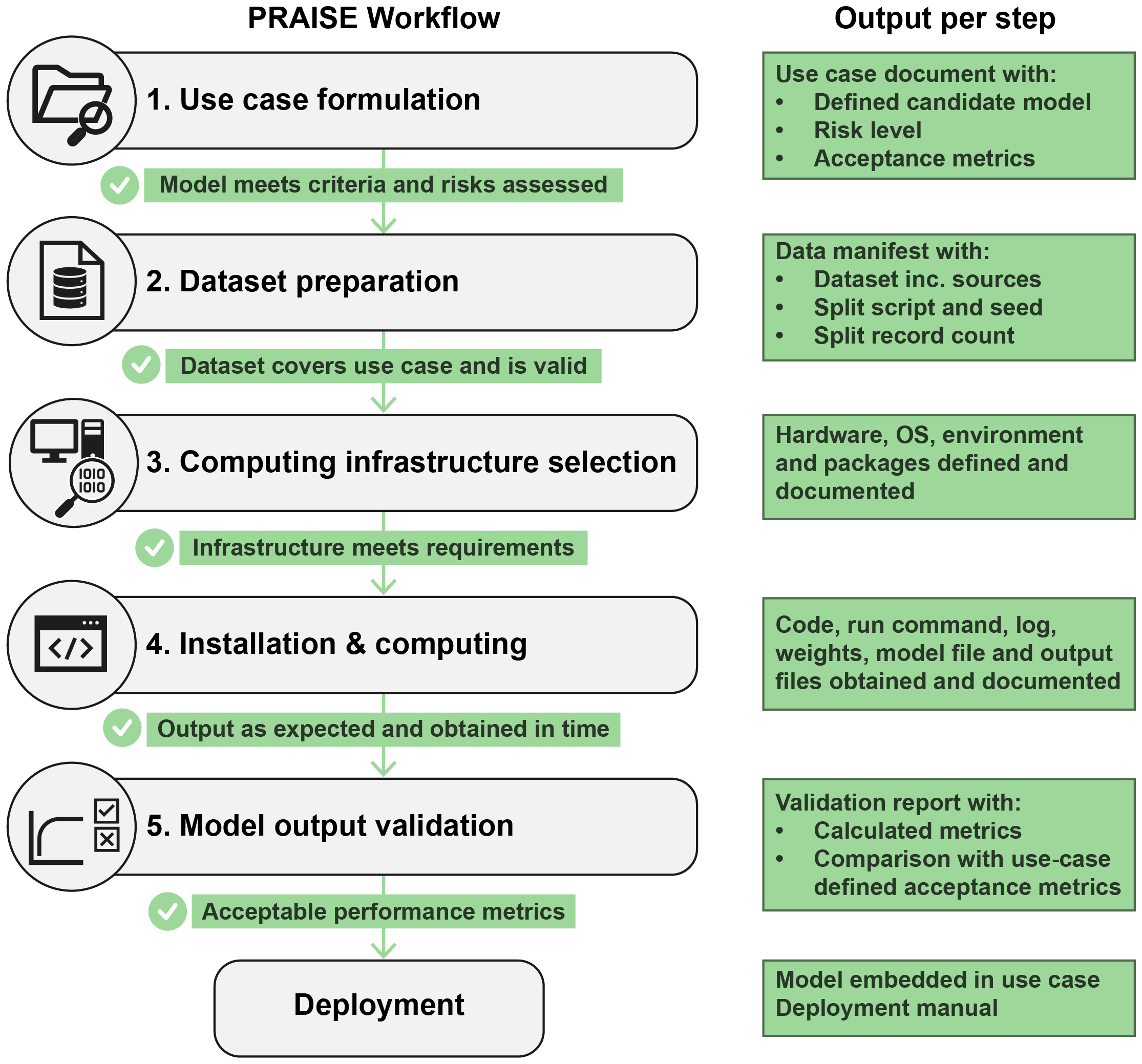}
  \caption{PRAISE (Practical, Reproducible AI-model Implementation for Science and Education) workflow for implementing scientific AI models to practical case studies, assuming no issues arise. Outputs and checks per step are given. Troubleshooting steps for common issues are shown in Section~\ref{sec:fallback-troubleshooting-by-workflow-iteration}.}
  \label{fig:praise}
\end{figure*}

\subsection{Environment and package managers}\label{sec:environment-and-package-managers}

AI models depend on a collection of software packages, often with strict version requirements. Environment and package managers (e.g., Conda, Mamba, Pip) facilitate the installation and isolation of these dependencies within dedicated environments. This isolation is critical, as incompatibilities between package versions or programming language versions are a common source of problems in AI workflows. Model developers frequently provide environment specifications (e.g., .yml files), which should be followed where possible.

\subsection{Development tools}\label{sec:development-tools}

Development tools offer an interface for writing, editing, and executing code. Examples of development tools are Visual Studio Code and JupyterLab. In this software, the application code can be opened and an environment selected. It is advisable to create the environment in an environment and package manager. Development tools are designed to be helpful to the human users, and often provide debugging features, file management and an integrated terminal.

\subsection{Generative AI as a coding aid}\label{sec:generative-ai-as-a-coding-aid}

Programming requires translating ideas into instructions a computer can execute, a skill that novices find difficult because of abstract concepts and entry barriers such as unfamiliar syntax, delimiters, indentation, and uninterpretable error messages \citep{Figueiredo2018}. When implementing AI models from literature however, domain expertise remains the primary requirement, while the coding barrier can be lowered through generative AI (GenAI) assistance. AI-assisted coding tools provide personalized help, on-demand feedback, and structured guidance in an educational setting \citep{Barzanji2025}, furthermore they have been shown to increase intrinsic motivation and reduce stress compared to unsupported approaches \citep{Fan2025}.

The underlying observations translate to non-programmer scientists. Access to AI coding support improves coding speed and quality without impairing the ability to understand or modify the resulting code \citep{Kazemitabaar2023}. This suggests that AI assistance can complement the development of practical programming skills. Across the whole programming workflow, these tools automate repetitive tasks, generate or explain code in natural language, and provide context-aware suggestions (e.g., relevant packages or functions) \citep{Treude2025,Sergeyuk2025}. Interaction modes range from conversational guidance to prompt-based code generation.

In practice, this support is accessed through tools such as GitHub Copilot, integrated into development tools like Visual Studio Code \citep{Zhou2025}, or assistants such as ChatGPT which can act as a collaborative programming partner \citep{Hamza2024}. These developments allow non-programmer scientists to engage with AI model implementation and rely less on extensive coding expertise.

\section{PRAISE: the workflow for the implementation of published AI models}\label{sec:praise-the-workflow-for-the-implementation-of-published-ai-models}

The Practical, Reproducible AI-model Implementation for Science and Education (PRAISE) workflow suggested in Figure~\ref{fig:praise} aims to help non-programmer scientists to implement existing scientific AI models in practical case studies. The expected output and checks the workflow should pass before continuation to the next step are given for each step. When these checks are not met, the workflow should be extended with fallback routes to address these issues. This section will purely focus on the basic workflow, and Section~\ref{sec:fallback-troubleshooting-by-workflow-iteration} will go into the fallback routes for potential issues.

Complete and accurate documentation should be maintained throughout the workflow, with each action recorded as it is performed to support later troubleshooting and reproducibility on other computer systems. A private Git repository is recommended as an effective way to maintain this documentation. Its version-control system records each modification chronologically, creating a traceable log of the workflow and supporting subsequent review, troubleshooting, and reproduction.

\subsection{Use case formulation}\label{sec:use-case-formulation}

Arguably the most important step in the PRAISE workflow is the proper formulation of a use case. A clearly defined problem is essential for determining whether AI or ML offers an appropriate solution, or whether conventional IT or data-management approaches would be more suitable. In short: AI/ML must address a defined problem rather than serve as a default solution. Moreover, this aids the selection of what kind of AI or ML is required.

\begin{figure}[tbp]
  \centering
  \includegraphics[width=0.9\columnwidth]{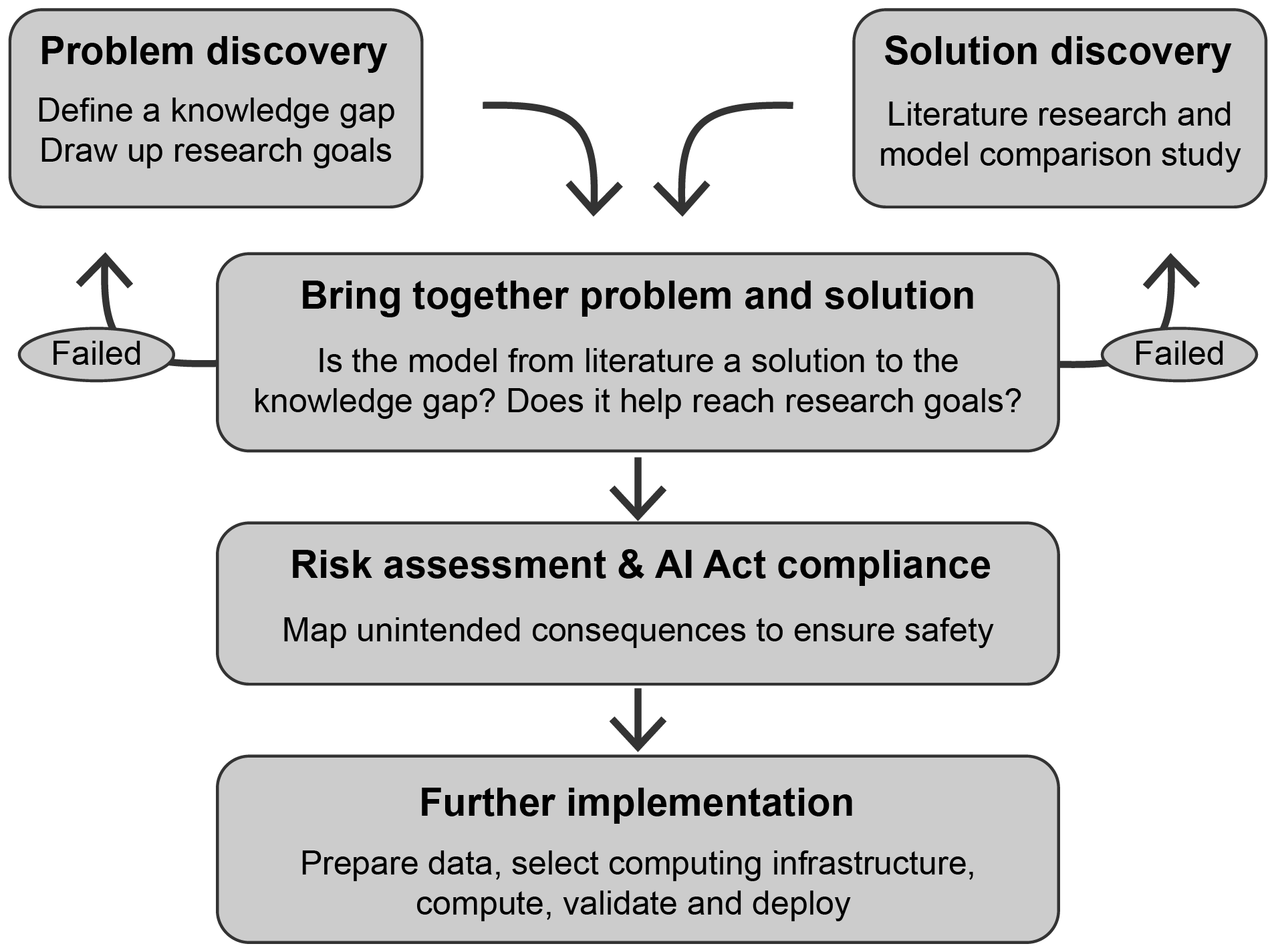}
  \caption{AI use case formulation guide. An AI use case can be constructed by bringing together both problem and solution discovery, until a candidate model is found that fits a defined knowledge gap. The regulatory compliance is performed afterwards.}
  \label{fig:usecase}
\end{figure}

Hofmann et al. proposed a guide to formulate AI use cases, as this is not a trivial undertaking \citep{Hofmann2020}. Here the proposed guide is followed with modifications in the reformulation of the `problem and solution alignment' and the inclusion of regulatory compliance (Figure~\ref{fig:usecase}).

\subsubsection{Simultaneous problem and solution discovery}\label{sec:simultaneous-problem-and-solution-discovery}

Central to the modification in problem and solution alignment from the original guide is that problem and solution discovery were performed consecutively, while in our view problem and solution discoveries can be performed simultaneously. This process should be repeated until the problem and the proposed solution are aligned. An initially vague problem can guide a search for potential AI solutions; studying these solutions clarifies which problems can be addressed, thereby enabling a more precise formulation of the original problem. This refined formulated problem is the starting point of the next iteration. The process naturally terminates when formulated problem and proposed AI solution become aligned.

When selecting potentially usable candidate models, the following should be considered:

\begin{itemize}
\item Recency: AI research advances fast, so a model of several years old is likely deprecated, although a more mature model often carries fewer bugs and more solved issues.
\item Developer support and community involvement: an active repository indicates that the model is kept working as dependencies and frameworks evolve, and that installation problems have been previously discussed.
\item Available validation options: models that ship with ways to assess accuracy, robustness or feature attribution are preferred, as adding these afterwards requires programming expertise (Section~\ref{sec:model-validation}). Acceptance values for the chosen validation metric (Section~\ref{sec:verifying-model-training-and-predictive-performance}) should be set here to determine when model performance is acceptable.
\end{itemize}

\subsubsection{Risk assessment and compliance}\label{sec:risk-assessment-and-compliance}

In addition, here the use case formulation guide by Hofmann et al.\ \citeyearpar{Hofmann2020} is extended by incorporating a risk assessment and compliance check, particularly in line with the EU AI Act \citep{EC2023}. This step is essential to incorporate data security and regulatory constraints from the start. This may include compliance with privacy regulations such as the General Data Protection Regulation (GDPR) \citep{EU2016}, intellectual property (IP) rights, and non-disclosure agreements (NDA). Information protection is strengthened by aligning local data handling with established standards such as ISO/IEC 27001 (information security management system) \citep{ISO2022a} and ISO/IEC 27701 (privacy information management system) \citep{ISO2025}. The safety of local large language model (LLM) use should be assessed against the Open Worldwide Application Security Project (OWASP) guidelines for LLM risks. These guidelines include (but are not limited to) prompt injection, sensitive information disclosure, and insecure output handling \citep{OWASP2024}. The use case reporting framework proposed by \citet{Hupont2024} can be used to systematically collect contextual information, in order to estimate the associated risk level of AI models. Particular caution is required when models process personal data or are applied in sensitive domains such as eligibility decisions, law, migration, or elections \citep{Hupont2024,vanVelzen2025}.

\subsubsection{Implementation and documentation}\label{sec:implementation-and-documentation}

Hofmann et al. (2020) describe implementation in terms of development responsibilities, system integration, data handling, and ethical considerations. The workflow presented here adds practical steps for implementing the model itself: dataset preparation, infrastructure selection, model training, and validation (Sections~\ref{sec:dataset-preparation} to \ref{sec:model-validation}), with troubleshooting in Section~\ref{sec:fallback-troubleshooting-by-workflow-iteration}. The output of use case formulation is a document that reports the defined problem, the intended use (users, input and output), the candidate model, the possible risk level, and the acceptance criteria for model performance. These criteria are set before training. They state which performance metrics (Section~\ref{sec:verifying-model-training-and-predictive-performance}) will be used to validate the model, for example on predictive performance, and which threshold values the model must reach to pass validation.

\subsection{Dataset preparation}\label{sec:dataset-preparation}

Data is the foundation of a predictive AI model and often the limiting factor in machine learning \citep{Roh2021}. Hence, the essential step of dataset preparation should be performed with care.

To be able to reproduce a model from literature, the dataset should be provided by the author of that model. Datasets should be Findable, Accessible, Interoperable and Reusable (FAIR) \citep{Wilkinson2016}. Often the used datasets are included with the source code at a repository maintained by the author.

Once access to the dataset has been obtained, its validity for the use case should be evaluated according to the criteria set out in Section~\ref{sec:data-as-the-basis-for-learning} such as: dataset size, training set gaps, biases or the absence of negative datapoints. Assessment should address how well the dataset represents expected situations in the intended use case and how its inevitable imperfections may affect model-training performance.

Depending on the outcome of these analyses, the dataset needs to be supplemented with additional data, to improve the coverage of the dataset. The required additional data may be gathered from public repositories, automated extraction of online sources (data scraping) or newly generated experimental or synthetic data. A manual literature search to build a dataset is time consuming. Data scraping models can speed up this process, although automatic data extraction from the internet must be performed responsibly \citep{Ayuso2024,Kahlon2026}.

Finally, as emphasized in Section~\ref{sec:data-as-the-basis-for-learning} the (improved) dataset should be divided into subsets for testing, training and validation, while avoiding data leakage as much as possible. The split can be checked on four points, together forming the leakage protocol, condensed from the leakage items proposed by Kapoor et al. \citep{Kapoor2024,Kapoor2023}.

\begin{itemize}
\item \emph{Split order.} The dataset is split into training, validation and test sets before any preprocessing, normalization, feature selection or augmentation. Preprocessing parameters are derived from the training set only.
\item \emph{Overlap.} After splitting, the subsets share no identical records. Where records are related or highly similar, they are assigned to the same subset according to a defined similarity criterion. The similarity criterion and strictness of the split are reported \citep{Kroll2024,Cui2025,Elmi2026}.
\item \emph{Use of the subsets}. The validation set is used for early stopping and hyperparameter choices only. The test set is used once, for the final metrics in Section~\ref{sec:model-validation}. Any part of the training pipeline that reads the validation or test set for another purpose fails this check \citep{Goodfellow2016b}.
\item \emph{Reproducibility.} The split is generated by a script with a fixed random seed \citep{Wegmeth2023}. Both are documented so the split can be regenerated.
\end{itemize}

The output of the dataset preparation is a data manifest, containing information of the sources of the data, record count per subset, and the script and seed for data splitting.

\subsection{Computing infrastructure selection}\label{sec:computing-infrastructure-selection}

The next step is to select the computing resources needed to run an AI model from literature. This involves choices regarding the properties of the computing resources, which may be local, cloud or hybrid (see Section~\ref{sec:computing-resources}). The selection of the computational resources is a delicate step in which prerequisites of the AI model and computing resources need to be balanced: if the computing facilities are too limited, the model might simply not run or take an inordinate timespan to execute, while if the resources are too large managing them becomes a daunting task and will possibly quickly exceed the available budget. The check for this step is to ensure the infrastructure meets the requirements. The output is a chosen infrastructure as detailed in Section~\ref{sec:computational-stack-and-development-tools-for-ai-models}.

\subsubsection{Constraints on computational infrastructure}\label{sec:constraints-on-computational-infrastructure}

The computing memory required to train and run AI models is often the hardware bottleneck. Memory requirements of scientific AI inferences are notoriously difficult to estimate and compare since they are dependent on multiple factors like parameter size, how intermediate embeddings are stored, and how detailed the input is \citep{Tian2025}. Even when considering these complications, the disparity in memory requirements between AI models is striking \citep{Tian2025}. In an AI model replication study, seven of the 30 models were not replicated because the specific hardware required was practically unobtainable \citep{Gundersen2025}.

The specifications of the computer infrastructure used by the model developer could be used as a guideline for a required computing memory. However, using a trained model often requires much less memory than the training procedure itself \citep{Sohoni2019}. Thus, if retraining an AI model is not part of the AI use case, and the trained model is provided, the user would not need to match the developer's computational setup. If retraining is part of the AI use case, matching the computing infrastructure is fair for an honest comparison and full reproducibility. When this is not possible due to the earlier described balance in resources, the training and computing time will be longer. If this is not acceptable, the model can be downsized or the step to cloud computing can be taken (Section~\ref{sec:revising-user-code}).

\subsubsection{Local hardware computing}\label{sec:local-hardware-computing}

Dedicated local hardware for computing is ideal if the available or intended hardware is close to the specifications of the infrastructure used by the author of the used AI model. Local computation resources have the advantage that training, running and adjusting AI models can be done without additional token costs and access to the facilities is unconstrained. However, the investment might be substantial if the hardware needs to be purchased for this purpose only.

\subsubsection{Cloud and hybrid computing}\label{sec:cloud-and-hybrid-computing}

Cloud computing is an option when local hardware cannot support the model, even after optimization. Compared to on-site infrastructure, cloud computing has relatively low startup costs and can be deployed quickly, but comes with higher costs for prolonged GPU usage \citep{Armbrust2010}. For small-scale or exploratory work, local infrastructure remains preferable when suitable hardware is available. It is also worthwhile to consider a hybrid approach where initial setup optimizations are done with the local hardware so that heavy computational jobs can be efficiently performed in the cloud infrastructure.

\subsubsection{OS selection}\label{sec:os-selection}

The basic rule of thumb for the choice of the operating system is to use the OS which was employed by the authors of the published AI model. Converting routines used on one platform to the next is in principle possible but is likely to fail given the complexity of using or training models and the many dependencies of the computational environment. If the required OS is not available, containers or full virtualizations can be used, though this often reduces performance. For local hardware, setting up a dual boot environment with different operating systems alongside each other might be an option.

\subsection{Running an AI model: from installation to computing}\label{sec:running-an-ai-model-from-installation-to-computing}

Running a published AI model requires its code and a matching software environment on the local machine. Training the model additionally requires the training dataset and more computational resources than inference with pretrained models.~Inference requires the pretrained model with its weights.

\subsubsection{Published AI model repository}\label{sec:published-ai-model-repository}

Requirements depend primarily on the architecture and implementation of the published AI model. Datasets, code and documentation of a published AI model are typically available on GitHub. The repository, i.e. all folders and files, can be cloned to the local machine through the git extension in a development tool like Visual Studio Code or JupyterLab.

\subsubsection{Development tools}\label{sec:development-tools-4-4-2}

The choice of development tools is highly personal. JupyterLab is web-based and designed to be user-friendly, Visual Studio Code needs to be installed and is more complex due to its customizability \citep{Kluyver2016}. Beginners in programming are advised to try multiple tools to determine the ones they may prefer. Open-source availability does not imply security; the code, dependencies, and documentation should therefore be reviewed critically.

\subsubsection{Computational environment}\label{sec:computational-environment}

Before executing the code, an environment should be configured (Section~\ref{sec:environment-and-package-managers}). The documentation accompanying the AI model specifies the required programming language version and packages. In the development tool, the appropriate environment can be selected and the required packages installed via a command in the integrated terminal. A check should be performed for all required files, like datasets or trained model weights. These often need to be downloaded separately. File paths in the code should be checked and possibly adjusted to fit the repository structure. GenAI tools like Codex, Claude Code or GitHub Copilot can be used as personal programming assistants (Section~\ref{sec:generative-ai-as-a-coding-aid}).

\subsubsection{Running the AI model}\label{sec:running-the-ai-model}

With the repository, environment and data locally available, the training or inference of the AI model can start. Ensure that the computational work is done by the GPU. Checkpoints can be saved in case the training stopped \citep{TensorFlow2024}. The output can be checked against expectation at each checkpoint. When the execution has been completed within an acceptable time, one can continue with validating the model (Section~\ref{sec:model-validation}). If the hardware appears to be insufficient for model execution, the model might be downsized, or hardware expanded (Section~\ref{sec:assessing-computational-infrastructure} and \ref{sec:revising-user-code}).

Once training is complete, the model is saved in framework-specific formats, for PyTorch or TensorFlow. Pretrained models are shared in these same formats. Hugging Face developed Safetensors, which saves metadata in JSON format and is universally usable for different programming languages. Local inference might then be possible, since these calculations require less computing power than training. Both locally trained and downloaded pretrained models can be loaded for inference on the local machine with dedicated code.

\subsubsection{Basic after-training checks}\label{sec:basic-after-training-checks}

Whether a training run completed as intended can be judged from the computing time, the presence of the expected output files, and the values these files contain. A computing time far below the expected value or missing output files indicates that the training did not complete. Output files in which every prediction has the same value, for example 0.5 for a binary classifier, indicate that the model did not learn during training. Validation is only useful once none of these signs are observed.

\subsection{Model validation}\label{sec:model-validation}

The basic question of the validation step is whether the results of the AI model warrant deployment. Reliable AI model predictions depend on the quality of the training data and on what the model has actually learned from it \citep{Hartung2025}. Weight distributions (Section~\ref{sec:verifying-model-training-and-predictive-performance}) can be compared between training runs to indicate whether each run produced a model of comparable quality, and a tool such as WeightWatcher automates this comparison by analyzing the weight matrix of every layer and reporting a quality value per layer, which helps to identify layers that remained undertrained or became overtrained \citep{Martin2021}.

The performance indicators introduced in Section~\ref{sec:verifying-model-training-and-predictive-performance} can be calculated on the validation set during training and on the test set once training is finished. Prediction scores and true labels are the input for ROC-AUC calculations. MCC operates on discrete class labels and therefore requires a threshold, commonly 0.5. Standard options (such as \texttt{roc\_auc\_score}, \texttt{average\_precision\_score}, \texttt{matthews\_corrcoef}) are available in scikit-learn \citep{Pedregosa2011}. Repeating the calculation over several random seeds or data splits indicates how stable the obtained values are. Whether the obtained predictive performance metrics are acceptable depends on the use case. Calculating the same indicators for the part of the test set that lies furthest from the training data shows how well the model covers gaps in the dataset and how it behaves on unseen input.

Performance indicators do not need to match those reported for the original model exactly. Variations in outcomes might be caused by differences in the random seed used for weight initialization, the order in which data are processed in parallel, or the GPU libraries (cuDNN and CUDA) \citep{Pham2021}. Training runs across three datasets and six AI networks, with a fixed random seed, resulted in accuracy variations of up to 2.9\% \citep{Pham2021}. Whether the validation was adequate is recorded in the use case formulation document (Section~\ref{sec:use-case-formulation}). If it does not meet these requirements, Section~\ref{sec:diagnosing-validation-issues} helps locate the cause.

\subsection{Deployment}\label{sec:deployment}

When validation meets the acceptance criteria set in the use case document (Section~\ref{sec:implementation-and-documentation}), the model can be deployed in the use case. Predictions for new input data can be made, in the context described in the use case document: who uses the model, which input they provide, and which output they need. A short test with new, representative input confirms that the model is ready for this use. The model should accept the input format of the users, return output they can interpret, and run on the hardware available to them. A low-code web-based user interface gives users without programming experience access to the trained model \citep{Kroll2023}.

At this point the logbook with documentation of the implementation process can be turned into a wiki that others can follow. An LLM such as Claude or Copilot is a useful drafting aid here, if the result is checked against the original implementation to avoid propagating errors. The outputs presented in Figure~\ref{fig:praise} collectively form the reproducibility package: use case document, data manifest and split script, hardware and OS record, environment file, run commands, logs, model files and metrics. This package meets the bronze standard of \citet{Heil2021}. Before publishing, the package can be checked against a reproducibility checklist such as REFORMS \citep{Kwak2025,Kapoor2024}. Code, data and documentation should be published in a public GitHub repository linked from the paper. Large datasets belong on Zenodo or DataverseNL, under their own DOI cited in the paper.

\section{Fallback troubleshooting by workflow iteration}\label{sec:fallback-troubleshooting-by-workflow-iteration}

\begin{figure*}[t]
  \centering
  \includegraphics[width=.75\textwidth]{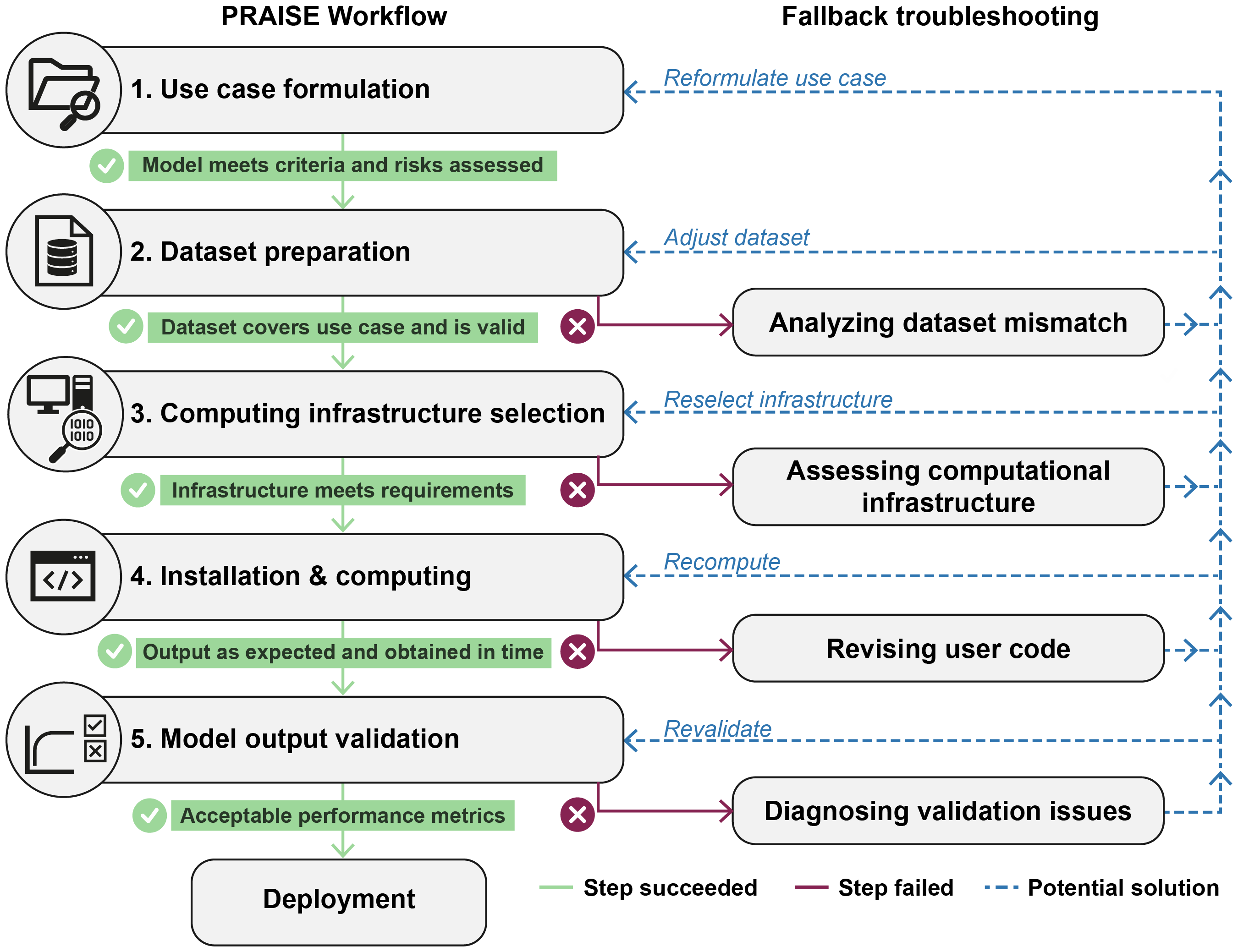}
  \caption{The PRAISE workflow for implementing scientific AI models to practical case studies, extended with troubleshooting routes. When a checkpoint is not met, the arrow marked with a cross leads to a decision point that helps decide which solution is required. These bring the user one or more steps back, depending on the issue. Such iterations are common in AI model implementations.}
  \label{fig:fallback}
\end{figure*}

The PRAISE workflow as presented in the previous section indicates the ideal situation, in which all steps can be executed without any difficulties. In practice, this will rarely be the case, thus the workflow needs to be adjusted accordingly at various stages when issues arise. The workflow diagram is extended with these troubleshooting routes (Figure~\ref{fig:fallback}).

\subsection{Analyzing dataset mismatch}\label{sec:analyzing-dataset-mismatch}

Dataset preparation fails when the use case and the available data do not match, because the dataset should cover the intended applicability domain of the model. The data needed for the use case might not be available or not of appropriate quality even after extension according to the procedures described in Section~\ref{sec:dataset-preparation}. The use case should then be reformulated to fit the available data while keeping the project relevant.

If changing the use case is undesirable and the mismatch comprises only minor gaps, dataset preparation and the remaining workflow can proceed, with the gaps documented as limitations on model validity. These limitations should be revisited during validation (Section~\ref{sec:model-validation}), where performance can be assessed separately across poorly represented regions of the applicability domain.

\subsection{Assessing computational infrastructure}\label{sec:assessing-computational-infrastructure}

Computational bottlenecks can sometimes be foreseen when the computing infrastructure is selected, but they can also become apparent during or after installation and test runs. In that case, the user code is revised first (Section~\ref{sec:revising-user-code}). The infrastructure is reassessed when code changes are insufficient. Infrastructure here covers all layers of Figure~\ref{fig:stack} below the user code: libraries, runtime, operating system and computing hardware.

It is essential to first analyze what the main underlying reasons for the issues could be. Common issues are that the system runs out of memory, runs out of storage capacity, takes too long to complete or delivers unexpected outputs. Possible remedies essentially fall into a few categories:

\begin{itemize}
\item adjust the computational infrastructure
\item adjust the datasets
\item reformulate the use case
\end{itemize}
Often a combination of these remedies is needed to solve computational infrastructure issues in the workflow.

\subsubsection{Reselect infrastructure}\label{sec:reselect-infrastructure}

The limits on computational performance can sometimes be raised with minor adjustments, such as extending the memory, acquiring more compute credits, or installing a different package in the environment. However, often this will entail a more major upgrade of the on-site hardware or move to cloud/hybrid computing (Section~\ref{sec:computing-resources}). To decide which adjustment is needed, the issues must be analyzed in detail to define the additional infrastructure requirements. A conclusion can be that the issues are not practically or financially solvable. In that case, the limitation must be accepted, and the solution shifts to reducing the dataset size or reformulating the use case (Section~\ref{sec:use-case-formulation} and Section~\ref{sec:dataset-preparation}).

\subsubsection{Adjust the dataset or reformulate the use case}\label{sec:adjust-the-dataset-or-reformulate-the-use-case}

With the experience gained to this point, a clearer picture of what questions and what kind of datasets the available computational infrastructure would be able to handle should be present. For example, the datasets might be reduced in size without losing the characteristics needed to use or train the model properly. Another option could be to modify the questions of the use case, such that they can be answered by the current setup or adjustments thereof.

\subsection{Revising user code}\label{sec:revising-user-code}

Problems during installation and computing, such as a run that crashes, runs out of memory, takes too long or gives unexpected outputs, can often be solved by revising the user code and recomputing. When the cause lies in computing infrastructure, for example a failed installation or package conflict, the infrastructure is reassessed (Section~\ref{sec:assessing-computational-infrastructure}).

\subsubsection{Model downsizing}\label{sec:model-downsizing}

To make a model fit the computational requirements within the existing infrastructure, downsizing of the model can be considered. One downsizing method is reducing the batch size, which is configurable in frameworks like PyTorch or TensorFlow, though smaller batches produce noisier gradients \citep{Keskar2017}. Other common approaches include gradient checkpointing, precision scaling, and gradient accumulation.

\subsubsection{Gradient checkpointing}\label{sec:gradient-checkpointing}

Another way to reduce infrastructure requirements is gradient checkpointing. This adjustment reduces memory usage by storing less information during the forward pass and recalculating intermediate values when needed \citep{Chen2016}. This adjustment reduced the memory consumption of a thousand-layer AI model sixfold with only a 30\% increase in computation time.

\subsubsection{Precision scaling}\label{sec:precision-scaling}

Further lowering of memory requirements can be achieved by precision scaling. This method reduces the numerical precision used within the model, for example from 32-bit to 16-bit values, thus resulting in lower memory requirements while often maintaining comparable model performance \citep{Dettmers2022}.

\subsubsection{Gradient accumulation}\label{sec:gradient-accumulation}

Gradient accumulation spreads gradient updates across multiple smaller batches before adjusting model parameters, allowing larger models to be trained on limited hardware \citep{Rajbhandari2020}. Regardless of the approach, the resulting model performance should be validated against outcomes reported in the literature to confirm that downsizing did not negatively affect results. If downsizing is insufficient, the use of pretrained models may be a more suitable alternative.

\subsubsection{Alternative types of machine learning}\label{sec:alternative-types-of-machine-learning}

As reviewed in Section~\ref{sec:from-input-data-to-a-prediction}, there are various types of machine learning with corresponding loss functions to be minimized. It is worth considering whether there are modifications in the applied training procedure to alleviate the computational bottlenecks (as different implementations might have impact on the required computational power, memory, etc.). If code changes do not resolve the issue, the next step is to reassess the infrastructure or to adjust the dataset or use case.

\subsection{Diagnosing validation issues}\label{sec:diagnosing-validation-issues}

If the validation of the trained model does not yield acceptable results, model diagnostics should be performed on the dataset, the scripts and/or notebooks. An unmodified reproduction of the published model is expected to reach performance metrics comparable to those reported by the authors, but variations in validation outcomes might occur (Section~\ref{sec:model-validation}).

\subsubsection{Dataset Composition and Validation Performance}\label{sec:dataset-composition-and-validation-performance}

Modifications to the dataset influence the validation outcome. In supervised learning, a different data split (Section~\ref{sec:data-as-the-basis-for-learning}) changes which samples the model is optimized on and which are held out, so the reported metrics shift as well \citep{Cawley2010}. Extending the original dataset with additional records can improve predictive performance and can widen the coverage of the model \citep{Rajput2023,Wang2026}.

\subsubsection{Permutation Testing for Data Leakage}\label{sec:permutation-testing-for-data-leakage}

When metrics are higher than expected, a permutation test can be performed. Here, the training labels are shuffled at random, and the model is retrained to check whether predictive performance is attributable to relationships in the original data. In the permutation test, performance should approach the random baseline (ROC-AUC $\approx$ 0.5 and MCC $\approx$ 0). Substantially higher performance indicates that information remains available to the model despite removal of the original label-feature relationship and may point to data leakage or an error in the training or evaluation procedure \citep{Ojala2010}.

\subsubsection{Diagnosing Deviations from the Published Model}\label{sec:diagnosing-deviations-from-the-published-model}

Finally, the training and evaluation scripts should be compared with the published implementation to identify unintended changes in data handling, preprocessing, model configuration, optimization, or evaluation. Changes to how data is preprocessed or separated before a training can introduce data leakage and result in misleading performance estimates \citep{Kapoor2024}. Other differences, such as the learning rate, batch size, number of training epochs, or model parameters, can alter the training outcome even when the dataset is unchanged. The diagnostics should therefore establish whether obtained unexpected validation performances originate from issues in the validation protocol, differences in the dataset or issues in the training process. If none of these solve the validation issues, the use case should be reformulated.

\section{Case study of workflow application in research and education}\label{sec:case-study-of-workflow-application-in-research-and-education}

\begin{figure*}[t]
  \centering
  \includegraphics[width=.75\textwidth]{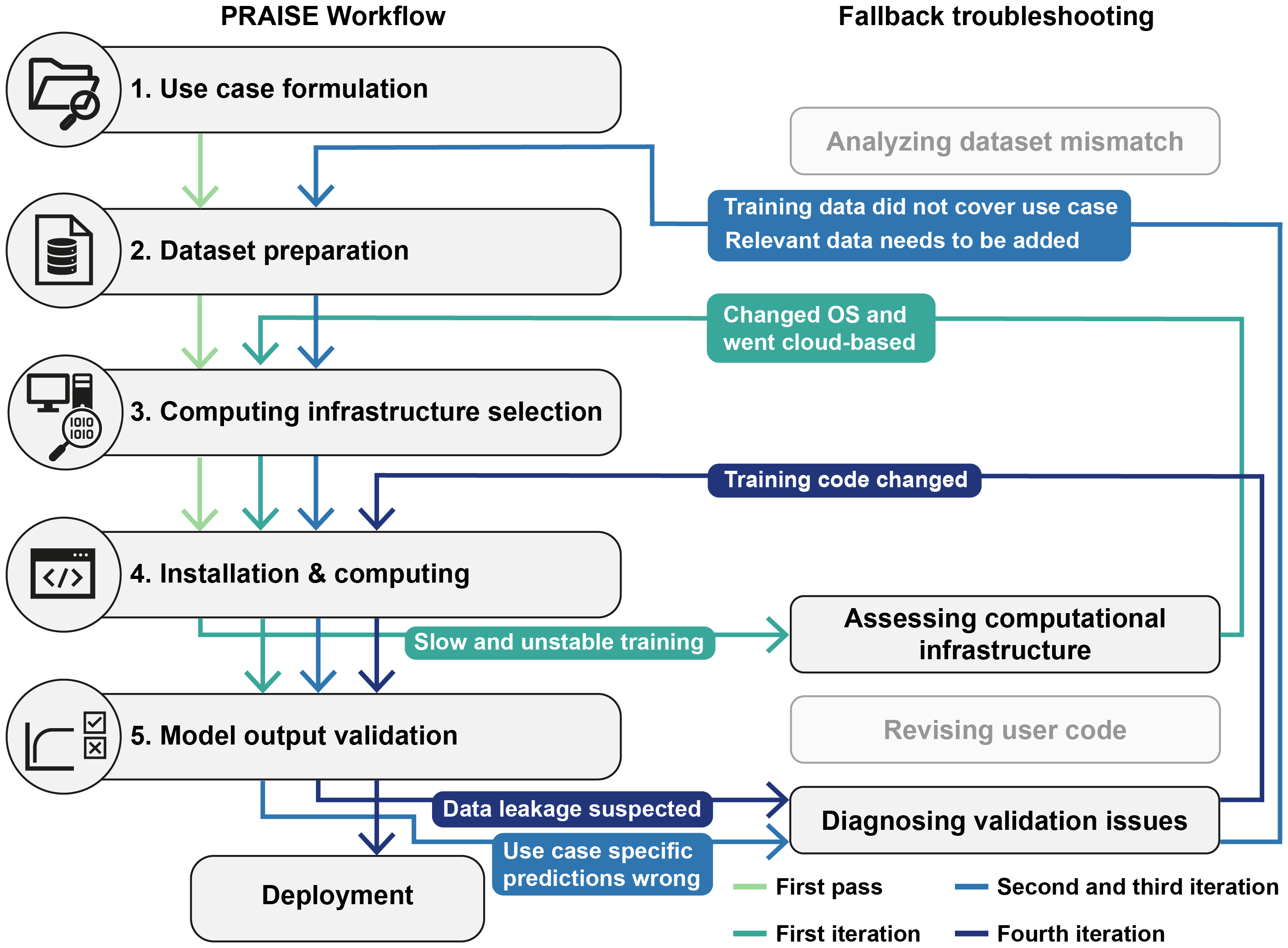}
  \caption{Implementation of an AI model to predict enzyme-substrate interactions following the PRAISE workflow. Colored arrows mark the first pass and the four iterations that followed, numbered in the order they occurred. First pass: the initial implementation of ProSmith, which stopped at the first issue. First iteration: the switch from Windows to a cloud-based Linux OS. Second iteration: the retraining on a dataset supplemented with data gathered manually from the literature. Third iteration: the retraining on an additional dataset containing true-negative data. Fourth iteration: the changes to the training pipeline to investigate suspected data leakage.}
  \label{fig:casestudy}
\end{figure*}

The preceding sections present the PRAISE workflow as a sequence of generic steps. This section illustrates how these steps were executed in a worked example. The sections follow the actual course of the project: a first pass through the steps of Figure~\ref{fig:fallback}, then four iterations, each triggered by an issue that required to return to a previous step (Figure~\ref{fig:casestudy}). These loops are an integral part of the PRAISE workflow.

The case concerns the prediction of enzyme-substrate interactions in a chemical recycling project. Three third-year BSc chemistry students and their tutor selected a published deep learning model, reproduced it, validated its performance and extended it to their own dataset within four months, using GenAI programming assistants. The tutor was a chemistry teacher and researcher, who helped formulate the use case and select computing infrastructure. None of them had a background in programming or data science.

\subsection{First pass: reproducing ProSmith}\label{sec:first-pass-reproducing-prosmith}

The first pass started with the formulation of the use case (Section~\ref{sec:use-case-formulation}). The work is part of C-Recycle, an EU Just Transition Fund project on chemical recycling, where waste streams such as plastic pyrolysis oil are upcycled into high-value chemicals that serve as feedstock for industry \citep{Stimulus2024}. Oxidative enzymes can functionalize these streams, but finding suitable enzyme--substrate pairs experimentally is laborious, and public databases only cover characterized pairs. The aim was therefore to reproduce an existing prediction model and deploy it in this research to match enzymes to substrates, find candidate enzymes and prioritize mutants for testing. The model would be considered ready for deployment when it reproduced the performance reported by its authors on the held-out test set and could predict interactions for new enzyme-substrate pairs.

With the use case defined, a limited literature search at the start of the project in 2024 identified the Enzyme Substrate Predictor (ESP) and subsequently ProSmith as suitable models \citetext{\citealp{Kroll2023}; \citealp{Kroll2024}}. Further models appeared during implementation and were also considered \citep{Du2025,Campbell2024,Elmi2025,Cui2025,Elmi2026}. Publication in this field is rapid enough that a selected model can be superseded before its implementation is completed, which makes model choice a recurring decision throughout the project, and possibly an iteration loop back to this step.

The risk assessment and compliance check of Section~\ref{sec:use-case-formulation} was carried out before implementation. No personal data were processed, and the application falls outside the sensitive domains identified there, placing the use case at a minimal risk level under the EU AI Act.

The project started with the ProSmith ESP dataset \citep{Kroll2024}. ProSmith was trained on a single NVIDIA A100 GPU with 40 GB memory, and training on a 24 GB NVIDIA RTX 6000 yielded similar performance at the cost of a longer run time. A 24 GB GPU would therefore suffice. This local training was done in the HR Datalab EAS, an in-house facility with high-performance workstations shared under clear usage guidelines. The workstations contained an NVIDIA GeForce RTX 4090 GPU with 24 GB VRAM and ran Windows 11. Windows was the operating system the students already knew, which lowered the entry barrier.

Installation guides in the programming community are typically written for Linux, and the ProSmith bash commands were incompatible with Windows PowerShell. Coding assistance (Section~\ref{sec:generative-ai-as-a-coding-aid}) was used to convert these commands to work on PowerShell, after which the original model could be reproduced.

\subsection{Iteration 1: unstable retraining on Windows}\label{sec:iteration-1-unstable-retraining-on-windows}

Retraining in step 4 of Figure~\ref{fig:casestudy} failed in multiple attempts, with errors that could not be traced to a specific cause. The operating system was identified as the likely cause, so implementation moved to a Linux environment on SURF Research Cloud, the national research computing infrastructure of SURF \citep{SURF2026a,SURF2026b}. The JupyterLab workspace ran Ubuntu 22.04 on a single NVIDIA A10 GPU. This was an iteration to step 3 of Figure~\ref{fig:casestudy}. After a fresh installation, baseline training ran without the earlier errors. Shorter training time was an added benefit: all training steps took less than two days, compared with about four days on the in-house workstations.

\subsection{Iteration 2: training data did not cover the use case}\label{sec:iteration-2-training-data-did-not-cover-the-use-case}

Validation of the predictions (step 5 in Figure~\ref{fig:casestudy}) showed that the enzymes and substrates of interest were absent from the training data. The project looped to the earlier step 2 in Figure~\ref{fig:casestudy}, an extended dataset was compiled from the literature. The ESP dataset was supplemented with enzyme-substrate pairs from online databases, each traced back to the original publication to verify its label. The larger dataset required targeted modifications to the preprocessing and gradient boosting modules.

\subsection{Iteration 3: simulated negatives in the training data}\label{sec:iteration-3-simulated-negatives-in-the-training-data}

Subsequent validation (step 5 in Figure~\ref{fig:casestudy}) showed that the training set contained simulated negatives, which can cause misclassification \citep{Campbell2024}. Negative interactions were scarce, because negative results are rarely published. The ProSmith ESP dataset compensates for this through random negative sampling. ProSmith was therefore trained on the VIPER dataset \citep{Campbell2024}, a high-throughput screening dataset with experimentally determined negative outcomes (step 2 in Figure~\ref{fig:casestudy}). The VIPER data were converted to the ProSmith input format, shuffled and split into test, training and validation sets using the ratio of \citet{Kroll2024}, with a fixed random seed for reproducibility.

\subsection{Iteration 4: suspected data leakage}\label{sec:iteration-4-suspected-data-leakage}

\begin{table*}[t]
  \centering
  \caption{Outcomes of five different ProSmith model training sessions and the corresponding validation scores. Each training was performed once, and the values are only indicative. The suspected data leak fix was investigated and found not to be data leakage.}
  \label{tab:metrics}
  \small
  \setlength{\tabcolsep}{4pt}
  \begin{tabularx}{\textwidth}{@{}>{\raggedright\arraybackslash}X>{\raggedright\arraybackslash}p{0.27\textwidth}cccc@{}}
    \toprule
    Model & Iteration & Accuracy & Precision & ROC-AUC & MCC \\
    \midrule
    Standard ProSmith Windows & Original & 0.93 & n.d. & 0.96 & 0.80 \\
    \addlinespace
    Standard ProSmith Linux & Install \& computing $\rightarrow$ infrastructure selection & 0.92 & 0.94 & 0.95 & 0.80 \\
    \addlinespace
    ProSmith manually supplemented database & Validation $\rightarrow$ data preparation & 0.91 & 0.92 & 0.95 & 0.76 \\
    \addlinespace
    ProSmith supplemented with VIPER database & Validation $\rightarrow$ data preparation & 0.93 & 0.83 & 0.95 & 0.75 \\
    \addlinespace
    Standard ProSmith suspected data leak fix & Validation $\rightarrow$ install \& computing & 0.90 & 0.92 & 0.95 & 0.74 \\
    \bottomrule
  \end{tabularx}
\end{table*}

The leakage protocol of Section~\ref{sec:dataset-preparation} indicated a potential issue since the validation set was used in training during the gradient boosting step. This came to light in step 5 in Figure~\ref{fig:casestudy}. The suspected leakage was reported to the developer \citep{Kroll2024issue}. Code inspection showed that the test set was used once and the ensemble weights were chosen on the validation set only, so check 3 of the leakage protocol was met. However, in the original pipeline the models were retrained on the training and validation data together before they were applied to the test set. Refitting the chosen model to all data used for selection is common practice \citep{Hastie2009}, but it means that the tested models were different from the validated models. For a reproducible pipeline, the students preferred a single set of models, trained on the training set only, which is validated, tested, and saved as one object. The cost of this choice is that the final models were trained on fewer records.

Two other adaptations were made. The trained gradient boosting models were discarded after training, and predictions were not automatically linked back to their original enzyme-substrate records. The pipeline was restructured to train the model on the training set only, to save models and ensemble weights automatically, and to map predictions to the original records (step 4 of Figure~\ref{fig:casestudy}).

\subsection{Comparison of performance metrics across training sessions}\label{sec:comparison-of-performance-metrics-across-training-sessions}

For each ProSmith training session, accuracy, precision, ROC-AUC and MCC (Section~\ref{sec:verifying-model-training-and-predictive-performance}) were calculated on the test set (Table~\ref{tab:metrics}). All trainings were executed once, so the values are indicative only. ProSmith remains predictive for the supplemented datasets, as the positive MCC scores show. The lower MCC after iteration 4 was suspected to be a result of data leakage, but was eventually deemed consistent with the smaller training set of the restructured pipeline and falls within the run-to-run variation reported for deep learning pipelines \citep{Pham2021}. The restructured pipeline met the criteria of the use case (Section~\ref{sec:first-pass-reproducing-prosmith}). Validation was therefore considered successful, deployment could commence.

\subsection{Deployment}\label{sec:deployment-6-7}

The retrained ProSmith model was deployed as a prediction script on SURF Research Cloud. The script loaded the saved model and ensemble weights, predicted interactions for new enzyme-substrate pairs, and mapped each prediction to its input record. The trained model could be used to screen enzymes for potential interactions with a substrate of interest, for example in functionalization reactions or in the degradation of waste streams. A second application was the prediction of how point mutations affected enzyme-substrate interaction. Both applications still require future experimental validation before the predictions can guide enzyme selection. The data, models and code of all iterations are available via the authors' GitHub page.

\subsection{Case study outcome}\label{sec:case-study-outcome}

Three BSc chemistry students and their tutor, all without programming experience, were able to reproduce and extend the predictive model ProSmith within four months using the PRAISE workflow. The model and computing infrastructure were selected by the institution, and the students performed the installation, data management, model training and validation independently. They adapted the pipeline and applied it to additional datasets (Table~\ref{tab:metrics}). The implementation required four iterations. Each started where the expected result was not obtained and the students returned to an earlier step (Figure~\ref{fig:casestudy}). The workflow allowed the students to understand each step and how to solve problems they encountered. A separate paper on the educational implementation of the PRAISE workflow into the BSc chemistry program is in preparation.

\section{Discussion}\label{sec:discussion}

We developed the structured, reproducible workflow PRAISE, which guides non-programmer scientists through implementing an AI model published in the literature. The workflow divides the process into explicit, verifiable steps and includes fallback loops for common issues.

By explicitly incorporating dataset and model validation, the PRAISE workflow positions reproducibility as a prerequisite in the implementation of AI models. The PRAISE workflow covers performance metrics that indicate whether predictions are accurate. The scientific value of the resulting predictions is a separate question that the user should assess themselves. Feature analysis can indicate which characteristics in the data are responsible for certain predictions \citep{Li2020}. LIME estimates feature importance for individual predictions by generating synthetic data around a single datapoint \citep{Linardatos2021,Ribeiro2016}. SHAP produces local explanations and global importance scores across the model \citep{Lundberg2017,Linardatos2021}. In models of the complexity of ProSmith, these methods are difficult to apply, because the model inputs are learned vector representations produced by several preprocessing steps, and individual dimensions carry no direct chemical interpretation. Feature analysis is therefore not included as a validation step in the proposed workflow.

Several other disclaimers for adopting the PRAISE workflow exist. One is that basic computer and data skills remain necessary, such as installing a program, understanding documentation and managing large datasets. In addition, the availability of the published model and its dataset, and the quality of the documentation, determine whether the workflow can be started at all. This becomes clear in the first step of the PRAISE workflow, where the use case is matched with potential AI solutions. Even with good documentation, specialized AI models may still require more advanced expertise than the workflow and generative AI coding helpers can give. If large AI models require more extensive computing power than the institution can provide, these models stay out of reach for users as well. The most restrictive limitation is the absence of data covering the research question of the use case, which results in unreliable predictions regardless of the quality of the model.

In conclusion, implementing an AI model from literature is a skill that can be learned. With the basic techniques described here and an understanding of the PRAISE workflow steps and their iterations, non-programmer scientists can reproduce published models and implement them in their own practice. A non-programmer scientist who adopts the PRAISE workflow becomes an evaluator and improver of AI models.

\section*{Authorship contribution statement}

Conceptualization: MA, SGN, VdB, SS, RvdW; Methodology: MA, SGN, RvdW; Software: MA, SS, ZY, PAS, AP, MvdW; Validation: MA, SS, ZY, PAS, AP; Investigation: MA, SS, ZY, PAS, AP, MvdW; Resources: SGN, VdB, SS, MvdW, RvdW; Data Curation: MA, SS, ZY, PAS, AP; Writing - Original Draft: MA, SGN; Writing - Review \& Editing: SGN, VdB, CM, RvdW; Supervision: MA, SGN, SS, CM, RvdW; Project administration: CM; Funding acquisition: CM

\section*{Data availability statement}

As the purpose of this paper is to present the PRAISE workflow and the specific use case, discussed in Section \ref{sec:case-study-of-workflow-application-in-research-and-education}, was included only to provide a concrete example of this workflow in practice, the results obtained in that use case do not constitute a publishable data for the present paper. In accordance with the practical suggestions of section \ref{sec:deployment}, the datasets, scripts, run commands, model files and validation results of the case study were stored in a private GitHub repository for internal use only. If the deployment of the use case leads to novel scientific insights, they will be made available with a future publication.
\section*{Declaration of competing interest}

The authors declare to have no known competing financial interests or personal relationships that could have influenced the reported work.

\section*{Declaration of generative AI and AI-assisted technologies in the manuscript preparation process}

During the preparation of this work the author(s) used Claude (Anthropic) to search for literature, interpret and write code, check grammar and shorten text. After using this tool, the authors reviewed and edited the content as needed and take full responsibility for the published article.

\section*{Funding}

This research was carried out within the C-Recycle consortium and co-funded by the European Union via the Just Transition Fund (JTF), project JTFZ-00101, administered by Stimulus Programmamanagement. The funder had no role in study design, data collection, analysis, or the decision to publish.

\section*{Acknowledgement}

The authors would like to thank the Rotterdam University of Applied Sciences Chemistry Department research group and the Engineering and Applied Sciences Datalab, as well as the C-Recycle Just Transition Fund consortium partners. This work used the Dutch national e-infrastructure with the support of the SURF Cooperative using grant no. EINF-17601.

\bibliography{references}

\end{document}